# JOINTLY MODELING ASPECT AND POLARITY FOR ASPECT-BASED SENTIMENT ANALYSIS IN PERSIAN REVIEWS

**Milad Vazan**, **Jafar Razmara\***

Department of Computer Science, Faculty of Mathematics, Statistics, and Computer Science,

University of Tabriz, Tabriz, Iran

**\***Corresponding author: razmara@tabrizu.ac.ir

## Abstract

Identification of user's opinions from natural language text has become an exciting field of research due to its growing applications in the real world. The research field is known as sentiment analysis and classification, where aspect category detection (ACD) and aspect category polarity (ACP) are two important sub-tasks of aspect-based sentiment analysis. The goal in ACD is to specify which aspect of the entity comes up in opinion while ACP aims to specify the polarity of each aspect category from the ACD task. The previous works mostly propose separate solutions for these two sub-tasks. This paper focuses on the ACD and ACP sub-tasks to solve both problems simultaneously. The proposed method carries out multi-label classification where four different deep models were employed and comparatively evaluated to examine their performance. A dataset of Persian reviews was collected from CinemaTicket website including 2200 samples from 14 categories. The developed models were evaluated using the collected dataset in terms of example-based and label-based metrics. The results indicate the high applicability and preference of the CNN and GRU models in comparison to LSTM and Bi-LSTM.

## 1. Introduction

Today, with the rapidly growing volume of user-generated text on the web, the interest in analyzing and understanding the users' opinions has arisen (Guellil et al. 2019). Sentiment Analysis (SA) is the computational investigation and identification of human tendencies and opinions expressed in textual documents (Korayem et al. 2016; Nandwani and Verma 2021; Singh et al. 2021). The SA techniques are extensively used in a variety of applications including text mining, web mining, and social media analytics to judge human behavior. The output of the analysis provides useful information for deciding different applications such as improving the quality of products based on the comments shared by their consumers. Regarding the large volume of textual data, it is highly important to employ fast and precise algorithms for automatically analyzing and extracting the opinions within these reviews. The design and development of such a solution need a deep knowledge of natural language processing, computational linguistics, and text mining (Habimana et al. 2020; Thet et al. 2010; Liu et al. 2019). Therefore, the SA systems enter the field to automatically classify text and analyze and identify users' opinions.

The problem of SA can be studied in three different levels including sentence level, document level, and aspect level (Thet et al. 2010). The sentence-level analyses the sentiment of an author in a sentence describing his/her polarity about an entity. At the document level, this analysis is done on a sequence of sentences whereas the whole text has been written based on polarity. At both sentence and document levels, when an entity receives a positive or negative sentiment based on a sentence or text, the sentiment does not cover all aspects of the entity. At the aspect level that is called Aspect-Based Sentiment Analysis (ABSA), a complete analysis of the sentiment about the existing entities is produced. The ABSA method investigates the opinions with different tendencies about each aspect in a sentence or text. Regarding the wide applications of ABSA, it has attracted tremendous attention in recent years and has become a leading methodology among the SA systems.

ABSA can be divided into two challenging tasks including Aspect-Term Sentiment Analysis (ATSA) and Aspect Category Sentiment Analysis (ACSA) (Zhu et al. 2019). ATSA focuses on identifying the sentiment of the term that occurs in the text and is divided into two sub-tasks namely Aspect Term Extraction (ATE) and Aspect Term Polarity (ATP) (Zeng et al. 2019). ATE tries to identify the aspect terms in a sentence while ATP is to determine the sentiment of each aspect term to be positive, negative, or neutral (Laskari and Sanampudi 2016). On the other hand, ACSA is divided into two important sub-tasks (Zeng et al. 2019) including Aspect Category Detection (ACD) and Aspect Category Polarity (ACP). Given a predefined set of aspect categories, the goal in ACD is to specify which aspect of the entity comes up in opinion, while ACP aims to specify the polarity of each aspect category from the ACD task.

The proposed solutions for ABSA can be grouped into two major strategies (Tao and Fang 2020): lexical-based techniques and machine learning-based techniques. Lexical-based techniques were the first in SA and work in an unsupervised manner making use of a predefined bag of words to extract polarity and sentiment in a text (Almatarneh and Gamallo 2019), while machine learning-based strategies rely on supervised classification of sentiments based on a set of labeled training samples (Tao and Fang 2020; Zhang et al. 2019). Machine learning-based techniques may employ traditional learning models such as maximum entropy (Mehra et al. 2002), naïve Bayes classifier (Malik and Kumar 2018), and support vector machines (Firmino et al. 2014), or may use deep learning models (Zhu and Qian 2018) which have the capability to provide more accurate results. These techniques use lexical features, sentiment lexicon-based features, parts of speech, or adjectives and adverbs as their input for analyzing sentiment, while their performance depends on the features. Recently, the use of deep learning-based methods in ABSA has received due attention because of their ability to automatically extract features from text (Zhu and Qian 2018). In this paper, a novel strategy is proposed to analyze sentiments in Persian reviews considering both ACD and ACP sub-tasks. Different deep learning-based models were examined to choose the best model in the structure of the final method. The proposed method was evaluated using a unique dataset of reviews on Persian movies. Following, after a review on the problem background and related works, we describe the details of the proposed method and the evaluation results.

**2. Related Works**
Regarding that the current study was conducted on both ACD and ACP problems, the following is a short review on related works on this area.

## 2.1 Aspect Category Detection

The previous works have mostly focused on solving the ACD problem considering it as a classification problem using supervised learning techniques such as support vector machines and neural networks. Movahedi et al. (2019) proposed a deep learning model based on the attention mechanism, which recognizes different categories by paying attention to different parts of a sentence. Ghadery et al. (2019) presented a supervised model called Language-Independent Category Detector (LICD), which is based on a text-matching technique without using any special tool or handmade feature extraction to recognize the aspect categories. The method assigns a sentence to a category if it has a high semantic similarity with the set of words representing a category. Otherwise, the sentence is assigned to a category if it is semantically and structurally similar to a sentence in a category. Xue et al. (2019) designed a model called Multi-Task neural Network for Aspect classification and extraction (MTNA) which concurrently solves two sub-tasks including aspect category classification and aspect term extraction. To this end, they employed an LSTM deep model to extract the aspect and a convolutional neural network (CNN) to classify text to perform the aspect classification. Tao and Fang (2020) proposed a model called Aspect Enhanced Sentiment Analysis (AESA), which uses multi-label classification to solve the ABSA for recognizing aspect categories and sentiment simultaneously. Ramezani et al. (2020) proposed a new contextual in-review representation using the BERT model which extracts useful features from text segments. Kumar et al. (2020) suggested a solution for the ACD problem based on a rule-based approach. Regarding the statistical limitation of association rules, they presented a hybrid rule-based approach that combines association rules with semantic associations and uses the concept of word embedding for semantic laws. Razavi and Asadpour (2017) proposed a new unsupervised approach for aspect recognition by employing a word embedding method to identify aspects using semantic words in the Persian language, called aspect keywords, and then, categorize aspects into semantic categories. To this end, the method first extracts the adjectives as words of opinion and uses them to construct the words of feeling. Frequent words are then extracted and selected as a list of candidate aspects. After pruning the candidate aspects using a set of exploratory rules, finally, the algorithm embeds the words based on the similarity criterion to the aspect and classifies them into categories.

## 2.2 Aspect Category Polarity

The main goal in ACP is to determine the polarity (positive, negative, or neutral) of the opinion related to each aspect. It is assumed that the aspects are already extracted in an initial step. The previous works mostly employ a supervised learning approach to solve the problem. Wang et al. (2016) employed an attention mechanism in two different deep learning models including AT-LSTM and ATAE-LSTM. The attention mechanism focuses on different parts of a text to identify different aspects in a text to analyze sentiment. To make better use of the aspect information, they used a vector for each aspect and appended it to the word input vector. Xue and Li (2016) designed a model based on a CNN model and a gating mechanism called GCAE. Their proposed model yielded more accurate and efficient results than LSTM. Their main idea was to use Tanh-ReLU units that can selectively produce a sentiment output depending on an aspect or entity. Due to the limitations in the preparation of labeled data in some applications, developing supervised learning approaches is not applicable. Thus, Fu et al. (2019) used a semi-supervised approach for aspect-level sentiment classification. To this end, they employed a variational autoencoder to manage unlabeled training data. Ma et al. (2017) proposed an Interactive Attention Network (IAN) model for aspect-level sentiment classification, which can logically pay attention to the words that are about the sentiment polarity of a specific aspect. Fan et al. (2018) proposed a convolutional memory network to learn both words and multiple words information as the attention mechanism taken from the convolutional operation of ideas. Chen et al. (2017) proposed a recurrent attention mechanism on memory to identify the sentiment of aspects. By adopting a multiple attention mechanism in this model, it has the ability to capture sentiment information separated by long distances. He et al. (2018) proposed an approach to improve the effectiveness of attention mechanism for aspect-level sentiment classification which can achieve a better achievement of representation of the aspect via capturing the semantic meaning of the given aspect. Huang et al. (2018) proposed an attention-over-attention (AOA) neural network to simultaneously learn the representation of aspects and sentences and can focus on the most important parts in both aspect and sentence automatically.

## 3. Background

### 3.1 Deep Learning

Recently, deep learning models have been developed successfully for different applications of natural language processing. Deep learning is the more powerful approach in the field of machine learning in which the features are selected automatically without human intervention through a multi-layered structure (Sorin et al. 2020). This is one of the major advantages of deep models in comparison to the traditional machine learning techniques where manual extraction of features or use feature selection algorithms (Dang et al. 2020). Among different models developed based on the deep learning approach, the following, five basic models are reviewed in summary.

### 3.2 Convolutional Neural Networks (CNNs)

CNNs constitute a special class of deep neural networks that have primarily developed to deal with two or three-dimensional data such as images. The structure of CNNs is highly similar to feed-forward neural networks with a specific spatial structure of convolution and pooling layers (Zhou et al. 2019; Kim 2014). Recently, one-dimensional models of CNN have been developed for processing patterns in different applications such as medical diagnosis (Salehi et al. 2021) and natural language processing (NLP) (Sorin et al. 2020). CNN has great power in extracting local features. Therefore, it performs well in tasks where feature identification is more important, such as sentiment analysis, because sentiments in the text are determined by key phrases (Yin et al. 2017). The model receives each pattern in the form of an input matrix, in which each row corresponds to a set of word vectors. The input pattern is passed through a number of convolution layers followed each by a pooling layer, and then, processed by a fully connected multilayer network to generate the final output. Figure 1 shows the framework of the CNN models in NLP applications.

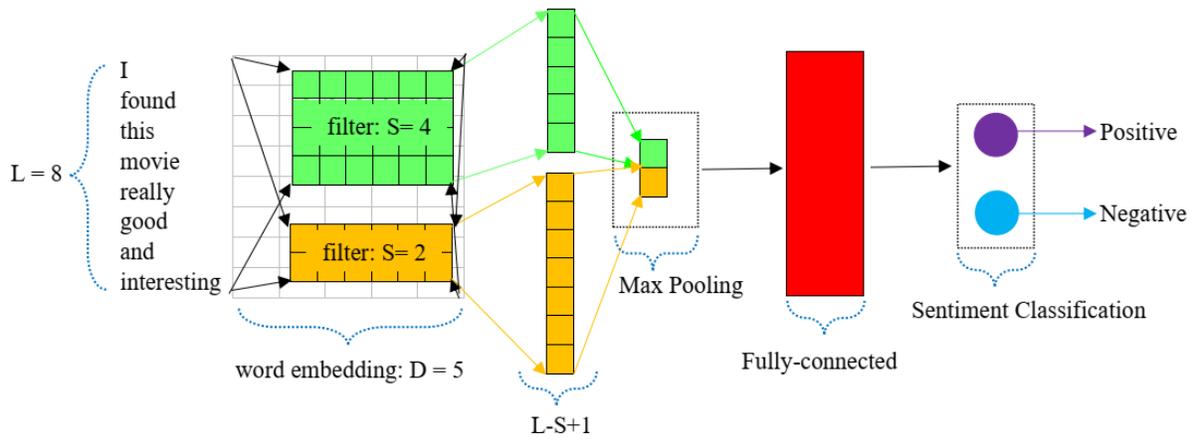

**Figure 1**. The architecture of a CNN model for sentiment classification

## 3.3 Recurrent Neural Networks (RNNs)

RNNs are a classical type of artificial neural networks designed to deal with data in sequence form. The structure of RNN is derived from feedforward neural networks and consists of state vectors in the hidden layer, which enable the model to memorize the previous input values (Zhou et al. 2019; Ramsundar et al. 2018; LeCun et al. 2015). In RNNs, the signals from the distant past are rapidly weakening. As a result, RNNs fail to learn models with complex dependencies on the distant past. RNNs show this problem in language modeling applications in which the meaning of a word depends on the words in the previous sentences (Ramsundar et al. 2018). Figure 2 represents the general architecture of RNNs.

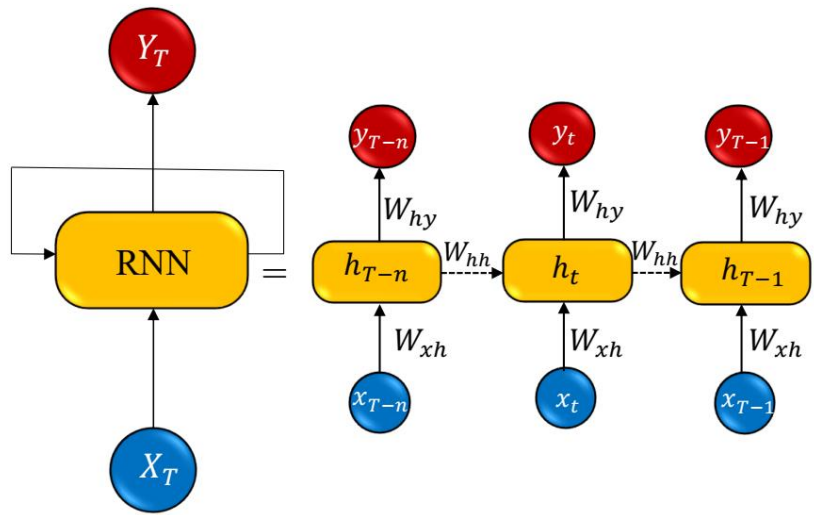

**Figure 2**. The general architecture of RNNs

## 3.4 Long Short-Term Memory (LSTM)

To solve the problem of long-term dependencies between the components of input patterns, LSTMs were introduced in 1997 by Hochreiter and Schmidhuber. LSTM networks are an extension of RNNs and differ in their ability to store information over a long period (Goodfellow et al. 2016). This is done by an important component in these networks called the state cell, the horizontal line running through the top of Figure 2, which determines whether new information should be added or deleted (Sorin et al. 2020).

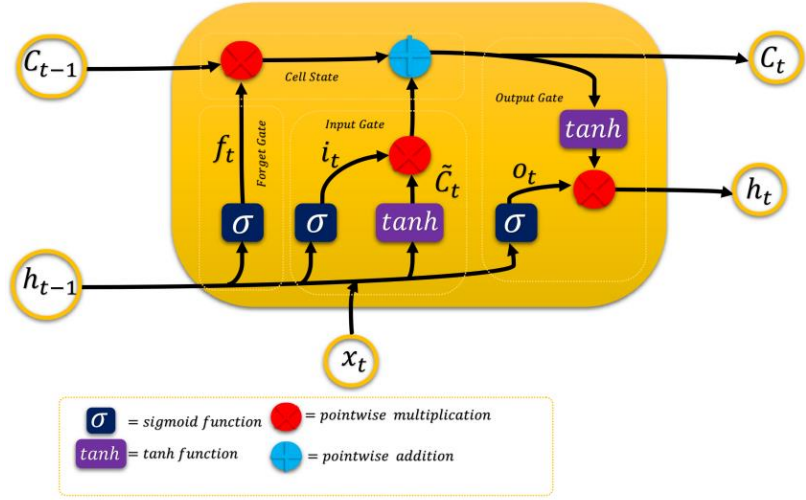

**Figure 3**. The architecture of LSTM

LSTMs add or remove information to the cell state called gates. An input gate ($i_t$), forget gate ($f_t$) and output gate ($o_t$) can be defined as (Alom et al. 2019):

$$f_t = \sigma(W_f \cdot [h_{t-1}, x_t] + b_f) \tag{1}$$

$$i_t = \sigma(W_i \cdot [h_{t-1}, x_t] + b_i) \tag{2}$$

$$\tilde{c}_t = tanh(W_c \cdot [h_{c-1}, x_t] + b_c) \tag{3}$$

$$o_t = \sigma(W_o \cdot [h_{t-1}, x_t] + b_o) \tag{4}$$

$$c_t = f_t * c_{t-1} + i_t * \tilde{c}_t \tag{5}$$

$$h_t = o_t * \tanh(c_t) \tag{6}$$

where $\sigma$ is the activation function, which normally is the sigmoid function. $W_f, W_i, W_o$, and $W_c$ are the weight matrices mapping the hidden layer input to the three gates and the input cell state. $b_f, b_i, b_c$, and $b_o$ are four bias vectors. $h_t$ and $h_{t-1}$ denote the current and previous hidden states, respectively. Also, $C_t$ is the cell state.

The input gate is responsible for protecting against unrelated entrances. The forge gate helps to forget past content, and the output gate plays a role in displaying or not displaying the contents of the memory cell (Patterson and Gibson 2017). The output of these gates depends on the previous hidden state and the current input (Zhou et al. 2019).

### 3.5 Gated Recurrent Unit (GRU)

The complexity of computations in LSTM has led researchers to simplify their equations, intending to simplify computations while maintaining the efficiency of the original model. In 2014, a modified RNN model was introduced called GRU. The GRU demonstrates a performance similar to, or sometimes better than, the LSTM with fewer parameters by excluding one LSTM member (Ramsundar et al. 2018; LeCun et al. 2015). GRU is now one of the most popular deep models to deal with data in sequence form. This popularity is due to its simplicity and low computational cost. Figure 4 shows the structure of GRU.

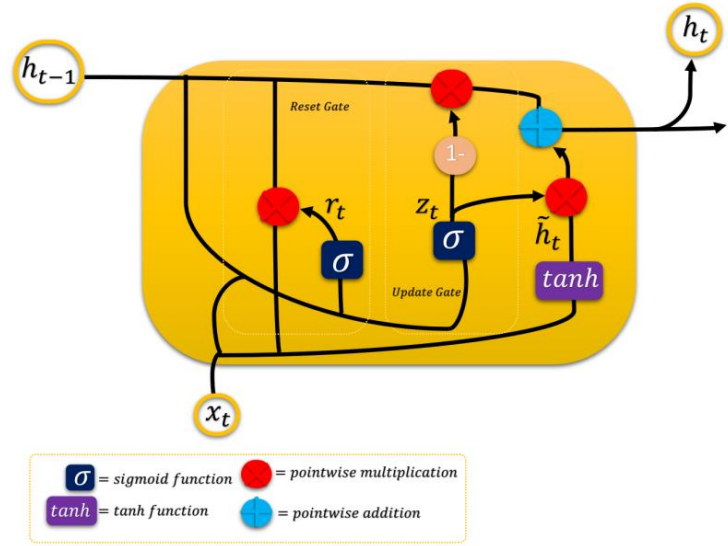

**Figure 4**. The general structure of GRU

Empirically, It has been observed that GRU, despite its simplicity compared to LSTM, has a performance close to LSTM. Mathematically the GRU can be expressed with the following equations (Zhou et al. 2019):

$$z_t = \sigma(W_z \cdot [h_{t-1}, x_t]) \tag{7}$$

$$r_t = \sigma(W_r \cdot [h_{t-1}, x_t]) \tag{8}$$

$$\tilde{h}_t = tanh(W \cdot [r_t * h_{t-1}, x_t]) \tag{9}$$

$$h_t = (1 - z_t) * h_{t-1} + z_t * \tilde{h}_t \tag{10}$$

where $W_z, U_z, W_o, b_z, W_r, U_r, b_r$ are the parameters of update and reset gates respectively. While $\tilde{h}$ and $h$ represent the intermediate memory and output respectively.

### 3.6 Bidirectional LSTM (bi-LSTM)

Bidirectional neural networks are designed to improve the performance of standard neural networks. In these networks, the input data can be entered into the network on both input and output sides through two different hidden layers. In bi-LSTM, a similar idea is used to process input patterns on both sides. The output of the forward layer is repeatedly calculated in a positive sequence of time, while the output of the backward layer is calculated in an inverse form. The Jointly Modeling Aspect and Polarity for Aspect based Sentiment Analysis in Persian Reviewsoutput in both front and rear layers is calculated similar to the conventional LSTM (Cui et al. 2020). The structure of bi-LSTM is represented in Figure 5.

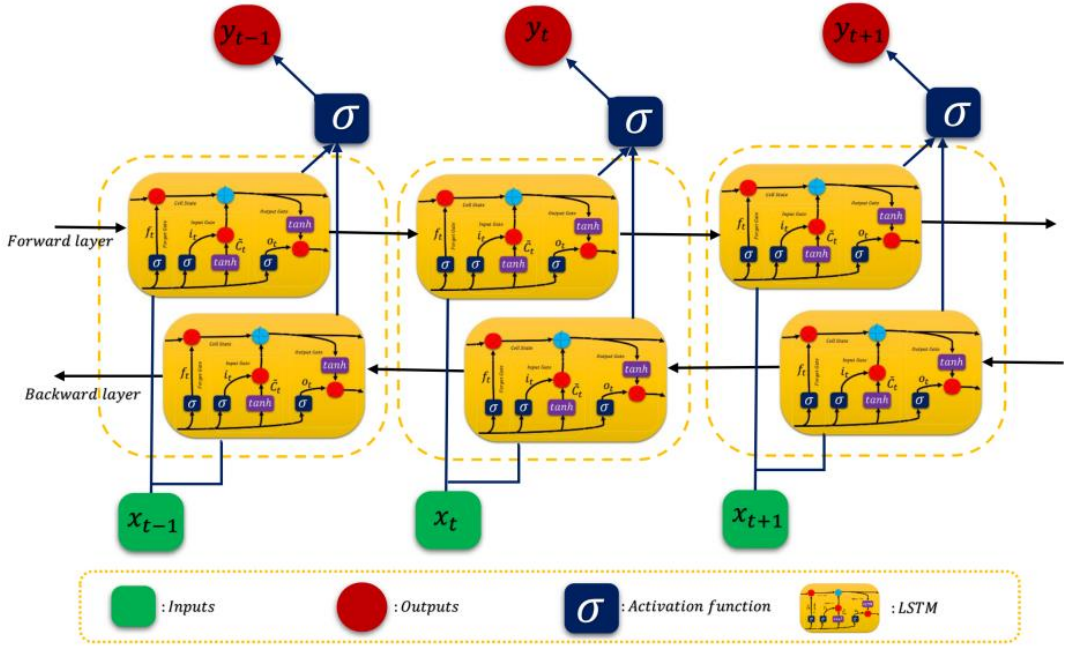

**Figure 5**. The general structure of bi-LSTM

## 4. Methods

The proposed approach to solve the ACD and ACP problems in Persian reviews is described in this section. The approach provides a multi-label classification, which can be used to solve two ACD and ACP sub-tasks, simultaneously. First, the formulation of these two problems into a problem is explained, and then, the architecture of deep models is described in detail.

### 4.1 Problem definition

Assume that a set of predefined categories $C = \{c_1, c_2, c_3, \dots, c_n\}$ and a set of reviews $R = \{r_1, r_2, r_3, \dots, r_m\}$ are available. The ACD task assigns each sentence in the set $R$ to one or more categories in the set $C$. Assume that the category $C$ for each sentence in the review dataset $R$ is predefined. Assume that the set $P = \{positive, negative\}$ is defined as the polarity of each aspect category. The ACP task assigns a polarity from the set $P$ for each sentence in the set $R$. Two ACD and ACP problems can be combined into a single problem by combining the input datasets. To this end, the Cartesian multiplication of two $C$ and $P$ sets is calculated as:

$$\boldsymbol{C \times P} = \{(\boldsymbol{c_1, positive}), (\boldsymbol{c_1, negative}), \dots, (\boldsymbol{c_n, positive}), (\boldsymbol{c_n, negative})\} \tag{11}$$

Whenever the classifier is used to predict the category of a sentence, it also recognizes the polarity of the sentence at the same time. Each sentence may be assigned into more than one category, and thus, the method carries out multi-label classification. Due to the possible structural similarity between the model sentences of both categories, a classifier may assign a sentence into both positive and negative categories. To prevent the model from producing such outputs, hereby, a simple idea is proposed called collision prevention thresholds (CPT). CPT is used to prevent conflicting categories from colliding based on predefined thresholds. To this end, the probability of two positive and negative categories is calculated and the polarity, which exceeds the predefined threshold, is selected.

### 4.2 The model architecture

The general architecture of the proposed model for solving both ACD and ACP problems is represented in Figure 6. Since each comment may be assigned into multiple categories, the problem is considered a multi-label classification task. As mentioned above, the polarity of each category can be positive or negative. In the output layer, the sigmoid activation function was used, as it generates a probability between zero and one for each category, independently. This output is converted to a binary value based on a threshold, which denotes the membership of the input to the corresponding category.

Before creating the model, the Keras Tokenizer API[1] was used to convert the input data into integers, and then, fed them to the embedding layers. The embedding layer receives three input parameters. These include the input size, which is the number of unique words, the length of the largest sentence, and the length of the embedding vector. The first two parameters are computed based on the data within the dataset while the third parameter, which is the length of the embedding vector, is an empirical hyperparameter (e.g. 32, 64, 128, 300, etc.) and in this work was set to 300. The dataset used in this study consists of 4730 unique words, and thus, an embedding layer with 4730 words, each with a vector size of 300, and the length of the largest sentence of 103 were created. The word embedding vectors are randomly set to floating-point values at first and then updated during the training phase of the model similar to optimization of the network weights.

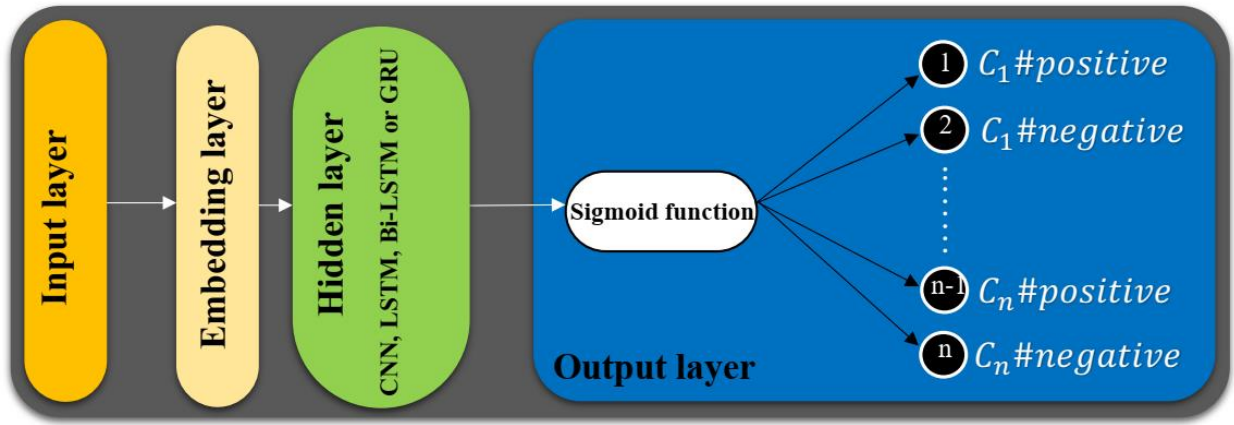

**Figure 6.** The general architecture of the proposed method

---

[1] https://keras.io/api/preprocessing/text/

Basic deep learning models have been employed in this study to evaluate their performance in ACD and ACP problems in comparison to other models. The models include CNN, LSTM, bi-LSTM, and GRU. The CNN model uses convolution operators to filter the input from the embedding layer. To this end, a number of 256 filters with size 3 were used to extract different features from the input. Then, a global max pooling is used to form the final features of the input pattern. Finally, a fully connected layer is used to process these features and make the final assignments into categories. For the LSTM, GRU, and bi-LSTM models, the input vector is given to the corresponding blocks of these models to extract textual information. In bi-LSTM, the input vectors can be given on both sides of the network. Finally, through a fully connected layer, this feature vector is mapped into the desired categories.

### 4.3 Training of the model

The essential step in the development of a deep model is network training. At this stage, the model tries to gain the ability to learn and predict desired outputs through optimizing its weights. Selecting an appropriate loss function, selecting an appropriate optimizer, preventing overfitting, and optimizing the hyper-parameters are the tasks that are carried out at this stage to build an efficient and practical model. In the following, these decision tasks are described in detail.

*The Loss function*

Selecting an appropriate loss function affects the convergence speed of a model. Cross-entropy is a popular loss function that is frequently used in discrete problems, especially in classification tasks. The function calculates the distance between two probability distributions and is defined as follows:

$$Cross\ Entropy(y, \hat{y}) = \frac{1}{n} \sum_{i=1}^{n} y_i \log(\hat{y}) \tag{12}$$

where $n$ is the number of training samples, $\hat{y}$ is the value predicted in the model and $y$ is the desired output. Binary cross-entropy is a typical form of the function that is used for binary classification and is defined as follows:

$$Binary\ Cross\ Entropy(y, \hat{y}) = -\frac{1}{n} \sum_{i=1}^{n} (y_i \log(\hat{y}) + (1 - y_i)(1 - \log(\hat{y}))) \tag{13}$$

*Weight optimization*

To update the network weights, the Nesterov-accelerated adaptive moment estimation optimizer (Nadam) was used. Nadam is a combination of Nesterov-accelerated gradient (NAG) and adaptive moment estimation (Adam) (Dozat 2016). Nadam has the advantage of fast convergence in comparison to other optimization algorithms for optimizing hyper-parameters.

*Overfitting prevention*

Overfitting occurs in the model when the model uses noisy data during the training phase. Two different methods were used to prevent overfitting occurrence. Batch normalization is the first method that prevents overfitting and improves the system performance through normalizing the input layer by adjusting and scaling (zero mean and unit variance) the input vector. Dropout is another low-cost but powerful computational method that was used in this study introduced by Hinton et al. (2012). Based on this method, each neuron remains with a probability $p$, is removed from the network with a probability of $1 - p$ in each epoch of the training phase. This strategy causes to ignore some features to be learned at each step and use different features for calculation of the output each time for this input data.

## 5 Results

### 5.1 Dataset

Due to the lack of a relevant dataset for ABSA in the Persian language, we have collected a set of user comments in the field of movie scoring from the CinemaTicket website (www.cinematicket.org). The dataset contains 14 different categories with a positive or negative polarity that have manually been labeled. Details of the collected review dataset are represented in Table 1.

**Table 1**. Details of the dataset collected from user's comments in the field of movie scoring from the CinemaTicket website (www.cinematicket.org)

| Category | positive | | negative | |
|---|---|---|---|---|
| | train | test | train | test |
| Actor | 152 | 11 | 67 | 6 |
| Acting | 200 | 17 | 110 | 7 |
| Story | 61 | 5 | 72 | 7 |
| dialogue | 7 | 3 | 36 | 3 |
| Style | 59 | 16 | 68 | 3 |
| the movie | 906 | 85 | 831 | 61 |
| screenplay | 61 | 7 | 111 | 8 |
| Filming | 35 | 3 | 21 | 1 |
| location | 14 | 3 | 6 | 1 |
| Content | 38 | 7 | 109 | 14 |
| Music | 22 | 3 | 7 | 1 |
| Issue | 56 | 9 | 69 | 6 |
| Director | 77 | 7 | 62 | 5 |
| Grimm | 8 | 1 | 4 | 2 |
| **total** | **1696** | **182** | **1573** | **125** |

### 5.2 Evaluation metrics

In this study, the ASBA problem is considered a multi-label classification, and thus, it is necessary to use appropriate evaluation metrics for this purpose. The evaluation process in multi-label classification is different and difficult than that of single-label classification (Zhang and Zhou 2013). Zhang and Zhoe (2013) divided the evaluation metrics for multi-label classification into two general categories including example-based and label-based as briefly reviewed following.

*Example-based Evaluation metrics*

These metrics calculate the performance of a classifier separately for each sample. Subset accuracy is a metric in this category that measures the number of correctly predicted labels based on the actual set. To this end, the fraction of correct predictions is calculated via the formula:

$$subsetAccuracy(h) = \frac{1}{n}\sum_{i=1}^{n}[h(x_i) == Y_i] \quad (14)$$

where $Y_i$ is the actual output for each sample $i$ and $h(x_i)$ is the classifier output. Hamming loss as another metric calculates the fraction of samples that are not labeled correctly using the formula:

$$hammingloss(h) = \frac{1}{n}\sum_{i=1}^{n}[h(x_i)\ XOR\ Y_i] \quad (15)$$

The metrics of accuracy, precision, recall and $F^\beta$ are more common metrics in classification tasks and can be used in multi-label classification using the following formula:

$$Accuracy_{exam}(h) = \frac{1}{n}\sum_{i=1}^{n}\frac{|h(x_i) \cap Y_i|}{|h(x_i) \cup Y_i|} \quad (16)$$

$$Precision_{exam}(h) = \frac{1}{n}\sum_{i=1}^{n}\frac{|h(x_i) \cap Y_i|}{|h(x_i)|} \quad (17)$$

$$Recall_{exam}(h) = \frac{1}{n}\sum_{i=1}^{n}\frac{|h(x_i) \cap Y_i|}{|Y_i|} \quad (18)$$

$$F^\beta{}_{exam}(h) = \frac{1}{n}\sum_{i=1}^{n}\frac{(1+\beta^2).Precision_{exam}(h).Recall_{exam}(h)}{\beta^2.Precision_{exam}(h) + Recall_{exam}(h)} \quad (19)$$

*Label-based Evaluation metrics*

These metrics are employed to measure the classifier performance separately for each category. Any binary classification metric (such as accuracy, precision, recall, etc.) can be used to evaluate the label-based metrics having True Positive ($TP_j$), True Negative ($TN_j$), False Positive ($FP_j$), and False Negative ($FN_j$) for each $yj$. Suppose that $B(TP_j, FP_j, TN_j, FN_j)$ denotes a specific binary classification metric where $B \in \{\text{Accuracy}, \text{Precision}, \text{Recall}, F^\beta\}$. Two kinds of label-based metrics including macro- and Micro -averaging are calculated as following (Zhang and Zhou 2013):

$$B_{Macro} = \frac{1}{q}\sum_{j=1}^{q}B(TP_j, FP_j, TN_j, FN_j) \quad (20)$$

$$B_{Micro} = B\left(\sum_{j=1}^{q} TP_j, \sum_{j=1}^{q} FP_j, \sum_{j=1}^{q} TN_j, \sum_{j=1}^{q} FN_j\right) \tag{21}$$

where *B* can be one of the following metrics:

$$Accuracy(TP_j, FP_j, TN_j, FN_j) = \frac{TP_j + TN_j}{TP_j + FP_j + TN_j + FN_j} \tag{22}$$

$$Precision(TP_j, FP_j, TN_j, FN_j) = \frac{TP_j}{TP_j + FP_j} \tag{23}$$

$$Recall(TP_j, FP_j, TN_j, FN_j) = \frac{TP_j}{TP_j + FN_j} \tag{24}$$

$$F^{\beta}(TP_j, FP_j, TN_j, FN_j) = \frac{(1+\beta^2).TP_j}{(1+\beta^2).TP_j + \beta^2.FN_j + FP_j} \tag{25}$$

### 5.4 Model hyper-parameters

The developed deep models use a set of hyper-parameters that are represented in Table 2. A simple manual adjustment method based on trial and error technique was used to adjust these hyper-parameters. To this end, different values for each hyper-parameter are selected and examined, the model output is evaluated in each step, and the best value is selected for each case. The value of hyper-parameters is represented in Table 2.

**Table 2**. The hyper-parameters of the model in this paper.

| Hyperparameter | Value |
| --- | --- |
| Number of hidden layer neurons | 200 |
| Dimensionality of word | 300 |
| Dropout | 0.5 |
| Epochs | 20 |
| Batch size | 50 |
| Maximum length | 103 |
| optimizer | Nadam |
| Learning rate | $1e^{-3}$ |
| kernel size | 3 |

### 5.5 Experimental Results

The proposed method was implemented using four different baseline deep models and evaluated based on both example-based and label-based metrics. Table 3 shows the evaluation results of the models based on example-based metrics for 50 independent execution with and without CPT. It can be seen from the table that the models obtain an increased subset accuracy and a decreased Hamming error when they use the CPT parameter. However, the models cannot perform well in terms of accuracy without using CPT. Furthermore, Table 4 shows the results obtained by the models for 50 independent execution based on the label-based metrics with and without CPT. It is obvious from the table that applying CPT enables the models to yield highly accurate results except for the GRU and bi-LSTM models.

Another study was conducted on the dataset to compare the efficiency of the models. This experiment was carried out using a 5-fold cross-validation scheme. Table 5 shows the average scores obtained by the models during ten executions with and without CPT based on example-based evaluation metrics during ten different execution. It is obvious from the table that the models with CPT obtain an increased subset accuracy. In terms of Hamming error, the models obtain approximately the same score with and without CPT. This is while the use of CPT did not work well and yield a slightly lower accuracy in other metrics. Furthermore, Table 6 shows the results obtained by the models during ten executions using 5-fold cross-validation based on label-based metrics. The results indicate that the use of CPT increases the performance of the models in terms of the recall metric, approximately the same average score in terms of accuracy, and a lower score in terms of precision and f1-score.

The diagrams drawn in Figures 7 to 9 show the different results of each model based on the example and label-based evaluation metrics. Also, these diagrams show a comparison between the performance of the models in the experimental data and the use of the 5-fold cross-validation method.

Table 3. Experimental results of the models on the test set for the example-based evaluation metrics

|  | Model | Subset. acc | Hamm. Loss | Acc. | Precision | Recall | $f_1$ |
|---|---|---|---|---|---|---|---|
| CNN | With CPT | 62.17 | 0.02412 | 74.43 | 78.92 | 79.95 | 79.43 |
| CNN | Without CPT | 59.66 | 0.02437 | 74.44 | 78.73 | 82.70 | 80.66 |
| LSTM | With CPT | 58.61 | 0.02607 | 70.36 | 76.32 | 74.30 | 75.29 |
| LSTM | Without CPT | 57.61 | 0.02614 | 70.41 | 76.27 | 75.50 | 75.87 |
| Bi-LSTM | With CPT | 59.38 | 0.02541 | 71.49 | 77.53 | 75.69 | 76.60 |
| Bi-LSTM | Without CPT | 58.42 | 0.02543 | 71.57 | 77.55 | 76.85 | 77.19 |
| GRU | With CPT | 62.10 | 0.0239 | 73.93 | 78.94 | 78.58 | 78.76 |
| GRU | Without CPT | 60.54 | 0.02405 | 73.95 | 78.81 | 80.28 | 79.53 |

Table 4. Experimental results of different models on the test set for the label-based evaluation metrics

|  | Model | Acc. | | Precision | | Recall | | $f_1$ | |
|---|---|---|---|---|---|---|---|---|---|
|  |  | macro | micro | macro | micro | macro | micro | macro | micro |
| GRU | With CPT | 97.59 | 97.59 | 56.92 | 76.26 | 65.07 | 78.42 | 57.24 | 77.32 |
| GRU | Without CPT | 97.56 | 97.56 | 63.71 | 76.71 | 59.90 | 78.72 | 58.36 | 77.70 |
| LSTM | With CPT | 97.39 | 97.39 | 49.21 | 70.70 | 62.17 | 78.79 | 51.55 | 74.49 |
| LSTM | Without CPT | 97.39 | 97.39 | 61.57 | 78.13 | 50.21 | 71.60 | 52.03 | 74.68 |
| Bi-LSTM | With CPT | 97.46 | 97.46 | 52.07 | 71.52 | 64.90 | 79.35 | 54.86 | 75.21 |
| Bi-LSTM | Without CPT | 97.46 | 97.46 | 64.24 | 78.74 | 53.00 | 72.43 | 55.20 | 75.43 |
| GRU | With CPT | 97.61 | 97.61 | 55.23 | 74.65 | 65.87 | 79.74 | 57.05 | 77.10 |
| GRU | Without CPT | 97.60 | 97.59 | 65.15 | 78.72 | 56.72 | 75.95 | 57.62 | 77.29 |

Table 5. Experimental results of the models on training set with 5-fold for example-based evaluation metrics

|  | Model | Subset. acc | Hamm. loss | Acc. | Precision | Recall | $f_1$ |
|---|---|---|---|---|---|---|---|
| CNN | With CPT | 58.96 | 0.02634 | 71.80 | 76.70 | 77.51 | 77.10 |
|  | Without CPT | 56.71 | 0.02679 | 71.66 | 76.36 | 79.73 | 78.00 |
| LSTM | With CPT | 52.31 | 0.03059 | 63.91 | 71.42 | 66.99 | 69.11 |
|  | Without CPT | 51.79 | 0.03053 | 64.04 | 71.54 | 67.75 | 69.56 |
| Bi-LSTM | With CPT | 56.51 | 0.02732 | 68.80 | 75.59 | 72.70 | 74.07 |
|  | Without CPT | 55.77 | 0.02735 | 68.89 | 75.53 | 73.68 | 74.58 |
| GRU | With CPT | 58.80 | 0.02581 | 71.47 | 77.28 | 76.15 | 76.71 |
|  | Without CPT | 57.43 | 0.02595 | 71.54 | 77.19 | 77.84 | 77.51 |

Table 6. Experimental results of the models on training set with 5-fold for label-based evaluation metrics

|  | Model | Acc. | | precision | | recall | | $f_1$ | |
|---|---|---|---|---|---|---|---|---|---|
|  |  | macro | micro | macro | micro | macro | micro | macro | micro |
| CNN | With CPT | 97.36 | 97.37 | 52.54 | 73.80 | 60.45 | 76.53 | 52.65 | 75.12 |
|  | Without CPT | 97.32 | 97.32 | 58.93 | 74.79 | 55.38 | 75.94 | 53.70 | 75.34 |
| LSTM | With CPT | 96.94 | 96.94 | 38.12 | 62.14 | 54.05 | 76.81 | 41.07 | 68.55 |
|  | Without CPT | 96.95 | 96.95 | 53.97 | 76.54 | 38.94 | 62.79 | 41.63 | 68.80 |
| Bi-LSTM | With CPT | 97.27 | 97.27 | 46.22 | 68.75 | 60.33 | 78.01 | 49.07 | 73.04 |
|  | Without CPT | 97.27 | 97.27 | 59.69 | 77.45 | 47.09 | 69.54 | 49.52 | 73.24 |
| GRU | With CPT | 97.42 | 97.42 | 51.42 | 72.32 | 62.11 | 78.18 | 53.06 | 75.12 |
|  | Without CPT | 97.40 | 97.40 | 61.57 | 77.16 | 52.83 | 73.68 | 53.79 | 75.37 |

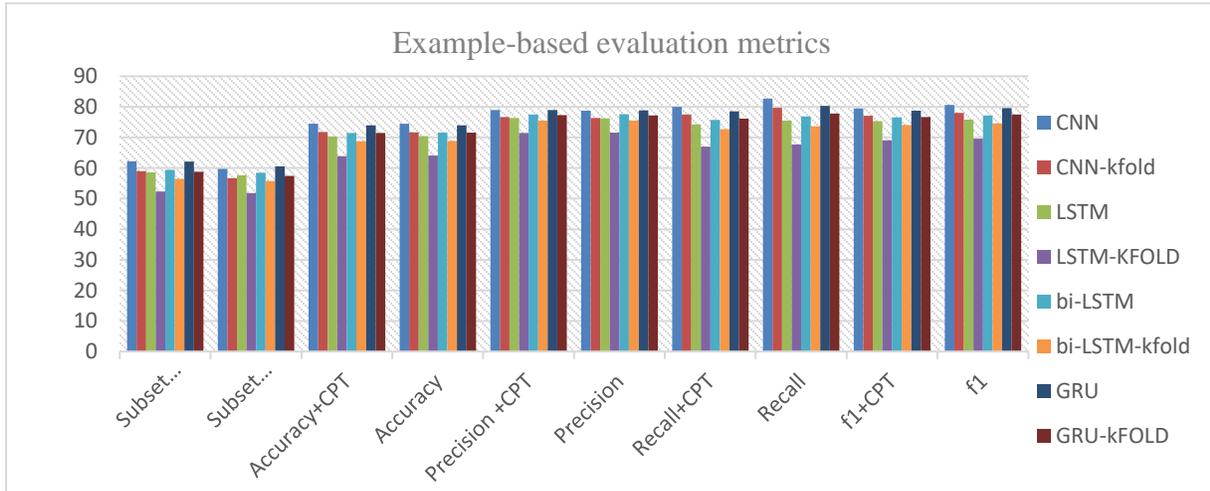

Figure 7. The results obtained by different models based on the example-based evaluation metrics

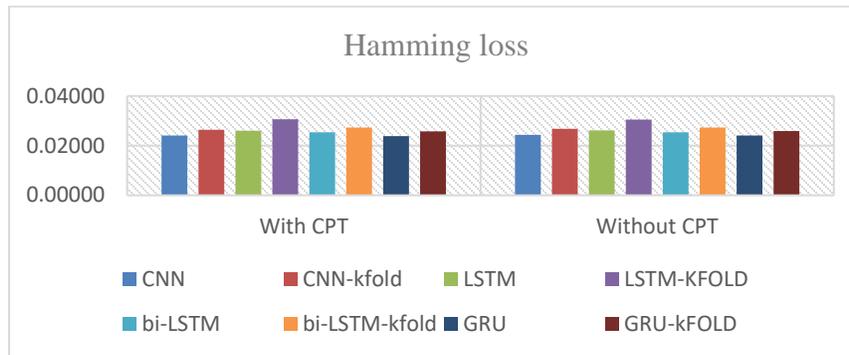

**Figure 8.** The results obtained by different models based on the Hamming Loss

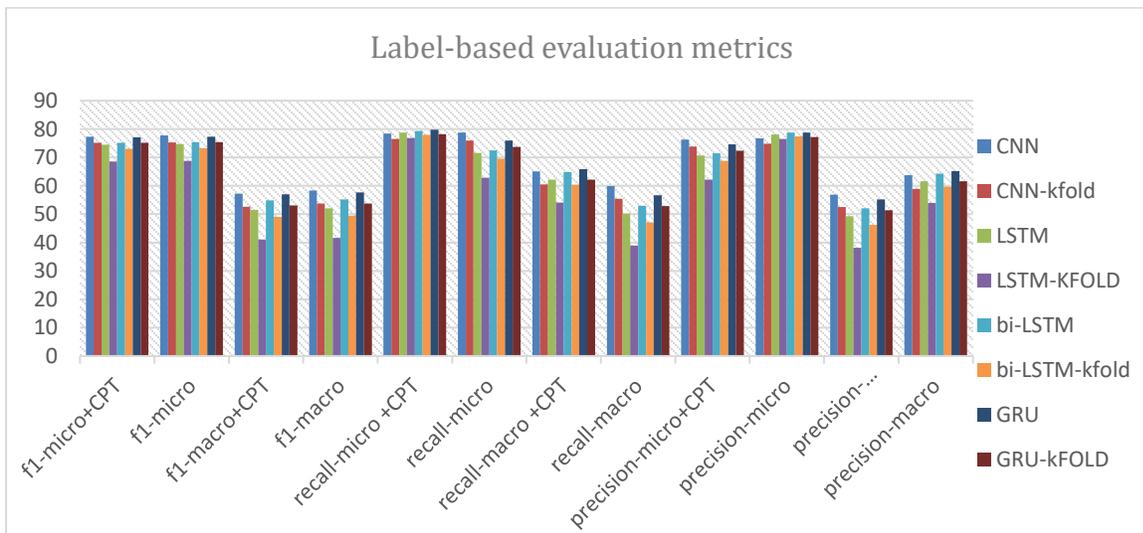

**Figure 9.** The results obtained by different models based on the label-based evaluation metrics

## 6. Discussion

Aspect-level sentiment analysis is an important issue in a variety of commercial applications. This is because all companies as well as users and customers tend to have a full and detailed understanding and analysis of merchandise, product, or service. In this regard, this research was conducted at the aspect level of sentiment analysis. At this level, different aspects of an entity are identified first, and then, the polarity is recognized for each identified aspect. In the previous works, these two tasks, i.e., identifying the aspect and determining the polarity, were considered as two different tasks and a model was developed separately for each task. Any incorrect identification of an aspect leads to failure in the appropriate analysis of its polarity. It is critically important to develop a model for both tasks to develop a highly accurate model.

In this study, we have developed a model that considers both tasks simultaneously in a single model for Persian comments. To this end, deep learning techniques were employed regarding their fruitful outcomes in the field of natural language processing. Deep learning models in natural language processing work almost similar to the models developed for image processing, where they try to identify relationships between words, sentences, and paragraphs instead of recognizing patterns from pixels. In this work, we have developed four basic deep models including CNN, LSTM, Bi-LSTM, and GRU. The input patterns are first preprocessed via the embedding layer, and then, entered into the deep models to generate the output. The output identifies the aspect and its polarity for each input pattern. Furthermore, to prevent the assignment of a sentence into both negative and positive aspects of a category, the CPT threshold was defined. The probability of both positive and negative categories is calculated and the polarity, which exceeds CPT, is selected.

The performance of the developed models was extensively evaluated based on both example-based and label-based metrics. Overall, it can be seen from the results in tables 3, 4, 5, and 6, that the CNN model and then GRU outperformed the rest of the deep models in processing both ACD and ACP data. This is because of the high performance of CNN in extracting local and fixed features. Regarding that emotions are expressed in sentences by specific words and phrases, CNN obtained a better outcome in finding these specific words. The point is also considered in the evaluation of RNN models. The GRU model gave better results in this study compared to the LSTM and bi-LSTM models. The reason for this can be related to the sentence structure of this data set as well as the type of problem. In addition, LSTM and bi-LSTM models have a better outcome in terms of recall in very long sentences. The sentences in this dataset are not very long and, as mentioned, specific words are decisive in the analysis of emotions, and the temporal dependence between words is not very important. Therefore, the GRU model was able to achieve better results due to its simpler structure and fewer learning parameters. Furthermore, the results indicate that considering the CPT threshold increases the accuracy of the models.

## 6. Conclusion

In this paper, two important sub-problems of aspect-based sentiment analysis, namely ACD and ACP, were considered simultaneously. To this end, the aspects and their polarities were grouped into different categories. Whenever the classifier receives an input pattern, it determines the category of the input regarding both the aspect and polarity of the pattern. Four different kinds of deep learning techniques were employed in the structure of the proposed method. The developed models were evaluated comparatively based on both example-based and label-based metrics. The results demonstrate that the CNN and GRU models outperform the rest deep models. In addition, the simpler structure of the GRU model besides its remarkable accuracy denotes the preference of this model in comparison to other deep models.

## References

Almatarneh S, Gamallo P. (2018). A lexicon based method to search for extreme opinions. PloS one 13(5):e0197816. https://doi.org/10.1371/journal.pone.0197816

Alom MZ, Taha TM, Yakopcic C, Westberg S, Sidike P, Nasrin MS, Hasan M, Van Essen BC, Awwal AA, Asari VK. (2019). A state-of-the-art survey on deep learning theory and architectures. Electronics 8(3):292.

Chen P, Sun Z, Bing L, Yang W. (2017). Recurrent attention network on memory for aspect sentiment analysis. In Proceedings of the 2017 conference on empirical methods in natural language processing, pp 452-461.

Cui Z, Ke R, Pu Z, Wang Y. (2020). Stacked bidirectional and unidirectional LSTM recurrent neural network for forecasting network-wide traffic state with missing values. Transportation Research Part C: Emerging Technologies 118:102674.

Dang NC, Moreno-García MN, De la Prieta F. (2020). Sentiment analysis based on deep learning: A comparative study. Electronics 9:483.

Dozat T. (2016). Incorporating Nesterov Momentum into Adam. In International Conference on Learning Representations Workshop.

Fan C, Gao Q, Du J, Gui L, Xu R, Wong KF. (2018). Convolution-based memory network for aspect-based sentiment analysis. InThe 41st International ACM SIGIR conference on research & development in information retrieval, pp 1161-1164.

Firmino Alves AL, Baptista CD, Firmino AA, Oliveira MG, Paiva AC. (2014). A Comparison of SVM versus naive-bayes techniques for sentiment analysis in tweets: a case study with the 2013 FIFA confederations cup. In Proceedings of the 20th Brazilian Symposium on Multimedia and the Web, pp 123-130.

Fu X, Wei Y, Xu F, Wang T, Lu Y, Li J, Huang JZ. (2019). Semi-supervised aspect-level sentiment classification model based on variational autoencoder. Knowledge-Based Systems 171:81-92.

Ghadery E, Movahedi S, Sabet MJ, Faili H, Shakery A. (2019). LICD: A language-independent approach for aspect category detection. In European Conference on Information Retrieval, pp 575-589. Springer, Cham. https://doi.org/10.1007/978-3-030-15712-8_37

Goodfellow I, Bengio Y, Courville A. (2016). Deep learning. MIT press.

Guellil I, Azouaou F, Mendoza M. (2019). Arabic sentiment analysis: studies, resources, and tools. Social Network Analysis and Mining 9(1):1-7. https://doi.org/10.1007/s13278-019-0602-x

Habimana O, Li Y, Li R, Gu X, Yu G. (2020). Sentiment analysis using deep learning approaches: an overview. Science China Information Sciences 63(1):1-36. https://doi.org/10.1007/s11432-018-9941-6

He R, Lee WS, Ng HT, Dahlmeier D. (2018). Effective attention modeling for aspect-level sentiment classification. In Proceedings of the 27th international conference on computational linguistics pp 1121-1131.

Hinton GE, Srivastava N, Krizhevsky A, Sutskever I, Salakhutdinov RR. (2012). Improving neural networks by preventing co-adaptation of feature detectors. arXiv preprint arXiv:1207.0580.

Hochreiter S, Schmidhuber J. (1997). Long short-term memory. Neural computation. 9(8):1735-80. https://doi.org/10.1162/neco.1997.9.8.1735.

Huang B, Ou Y, Carley KM. (2018). Aspect level sentiment classification with attention-over-attention neural networks. InInternational Conference on Social Computing, Behavioral-Cultural Modeling and Prediction and Behavior Representation in Modeling and Simulation, pp. 197-206. Springer, Cham. https://doi.org/10.1007/978-3-319-93372-6_22

Kim Y. (2014). Convolutional Neural Networks for Sentence Classification. arXiv preprint arXiv:1408.5882.

Korayem M, Aljadda K, Crandall D. (2016) Sentiment/subjectivity analysis survey for languages other than English. Social network analysis and mining 6(1):1-7. https://doi.org/10.1007/s13278-016-0381-6

Kumar A, Saini M, Sharan A. (2020). Aspect category detection using statistical and semantic association. Computational Intelligence 36(3):1161-82.

Laskari NK, Sanampudi SK. (2016). Aspect based sentiment analysis survey. IOSR Journal of Computer Engineering (IOSR-JCE) 18(2):24-8.

LeCun Y, Bengio Y, Hinton G. (2015). Deep learning. nature 521:436-44. https://doi.org/10.1038/nature14539.

Liu R, Shi Y, Ji C, Jia M. (2019). A survey of sentiment analysis based on transfer learning. IEEE Access 7:85401-12.

Ma D, Li S, Zhang X, Wang H. (2017). Interactive attention networks for aspect-level sentiment classification. arXiv preprint arXiv:1709.00893.

Malik V, Kumar A. (2018). Sentiment Analysis of Twitter Data Using Naive Bayes Algorithm. International Journal on Recent and Innovation Trends in Computing and Communication 6(4):120-5.

Mehra N, Khandelwal S, Patel P. (2002). Sentiment identification using maximum entropy analysis of movie reviews. Stanford University, USA.

Movahedi S, Ghadery E, Faili H, Shakery A. (2019). Aspect category detection via topic-attention network. arXiv preprint arXiv:1901.01183.

Nandwani P, Verma R. (2021). A review on sentiment analysis and emotion detection from text. Social Network Analysis and Mining 11(1):1-9. https://doi.org/10.1007/s13278-021-00776-6

Patterson J, Gibson A. (2017). Deep learning: A practitioner's approach. O'Reilly Media, Inc.

Ramezani S, Rahimi R, Allan J. (2020). Aspect Category Detection in Product Reviews using Contextual Representation. In Proceedings of ACM SIGIR Workshop on eCommerce (SIGIR eCom'20).

Ramsundar B, Zadeh RB. (2018) TensorFlow for deep learning: from linear regression to reinforcement learning. O'Reilly Media, Inc.

Razavi SA, Asadpour M. (2017). Word embedding-based approach to aspect detection for aspect-based summarization of persian customer reviews. In Proceedings of the 1st International Conference on Internet of Things and Machine Learning, pp 1-10. https://doi.org/10.1145/3109761.3158403

Salehi M, Razmara J, Lotfi S, Mahan F. (2021). A One-Dimensional Probabilistic Convolutional Neural Network for Prediction of Breast Cancer Survivability. The Computer Journal. https://doi.org/10.1093/comjnl/bxab096.

Singh M, Jakhar AK, Pandey S. (2021). Sentiment analysis on the impact of coronavirus in social life using the BERT model. Social Network Analysis and Mining 11(1):1-1. https://doi.org/10.1007/s13278-021-00737-z

Sorin V, Barash Y, Konen E, Klang E. (2020). Deep learning for natural language processing in radiology—fundamentals and a systematic review. Journal of the American College of Radiology 17(5):639-48.

Tao J, Fang X. (2020). Toward multi-label sentiment analysis: a transfer learning based approach. Journal of Big Data 7(1):1-26. https://doi.org/10.1186/s40537-019-0278-0

Thet TT, Na JC, Khoo CS. (2010). Aspect-based sentiment analysis of movie reviews on discussion boards. Journal of information science 36:823-48. https://doi.org/10.1177/0165551510388123

Vasilev I, Slater D, Spacagna G, Roelants P, Zocca V. (2019). Python Deep Learning: Exploring deep learning techniques and neural network architectures with Pytorch, Keras, and TensorFlow. Packt Publishing Ltd.

Wang Y, Huang M, Zhu X, Zhao L. (2016). Attention-based LSTM for aspect-level sentiment classification. In Proceedings of the 2016 conference on empirical methods in natural language processing, pp 606-615.

Xue W, Li T. (2018). Aspect Based Sentiment Analysis with Gated Convolutional Networks. In Proceedings of the 56th Annual Meeting of the Association for Computational Linguistics (Volume 1: Long Papers), pp 2514-2523.

Xue W, Zhou W, Li T, Wang Q. (2017). MTNA: a neural multi-task model for aspect category classification and aspect term extraction on restaurant reviews. In Proceedings of the Eighth International Joint Conference on Natural Language Processing (Volume 2: Short Papers), pp 151-156.

Yin W, Kann K, Yu M, Schütze H. (2017). Comparative study of CNN and RNN for natural language processing. arXiv preprint arXiv:1702.01923.

Zeng Z, Ma J, Chen M, Li X. (2019). Joint learning for aspect category detection and sentiment analysis in chinese reviews. In China Conference on Information Retrieval, pp 108-120. Springer, Cham. https://doi.org/10.1007/978-3-030-31624-2_9

Zhang C, Li Q, Song D. (2019). Syntax-aware aspect-level sentiment classification with proximity-weighted convolution network. In Proceedings of the 42nd International ACM SIGIR Conference on Research and Development in Information Retrieval, pp 1145-1148.

Zhang ML, Zhou ZH. (2013). A review on multi-label learning algorithms. IEEE transactions on knowledge and data engineering 26(8):1819-37.

Zhou J, Huang JX, Chen Q, Hu QV, Wang T, He L. (2019). Deep learning for aspect-level sentiment classification: survey, vision, and challenges. IEEE access 7:78454-83.

Zhu P, Chen Z, Zheng H, Qian T. (2019). Aspect aware learning for aspect category sentiment analysis. ACM Transactions on Knowledge Discovery from Data (TKDD) 13(6):1-21. https://doi.org/10.1145/3350487

Zhu P, Qian T. (2018). Enhanced aspect level sentiment classification with auxiliary memory. In Proceedings of the 27th International Conference on Computational Linguistics, pp 1077-1087.